\PassOptionsToPackage{unicode}{hyperref}
\PassOptionsToPackage{hyphens}{url}
\PassOptionsToPackage{dvipsnames,svgnames,x11names}{xcolor}
\documentclass[
doublespace,
  times]{anzsauth}

\usepackage{amsmath,amssymb}
\usepackage{iftex}
\ifPDFTeX
  \usepackage[T1]{fontenc}
  \usepackage[utf8]{inputenc}
  \usepackage{textcomp} 
\else 
  \usepackage{unicode-math}
  \defaultfontfeatures{Scale=MatchLowercase}
  \defaultfontfeatures[\rmfamily]{Ligatures=TeX,Scale=1}
\fi
\usepackage{lmodern}
\ifPDFTeX\else
\fi
\IfFileExists{upquote.sty}{\usepackage{upquote}}{}
\IfFileExists{microtype.sty}{
  \usepackage[]{microtype}
  \UseMicrotypeSet[protrusion]{basicmath} 
}{}
\makeatletter
\@ifundefined{KOMAClassName}{
  \IfFileExists{parskip.sty}{%
    \usepackage{parskip}
  }{
    \setlength{\parindent}{0pt}
    \setlength{\parskip}{6pt plus 2pt minus 1pt}}
}{
  \KOMAoptions{parskip=half}}
\makeatother
\usepackage{xcolor}
\makeatletter
\ifx\paragraph\undefined\else
  \let\oldparagraph\paragraph
  \renewcommand{\paragraph}{
    \@ifstar
      \xxxParagraphStar
      \xxxParagraphNoStar
  }
  \newcommand{\xxxParagraphStar}[1]{\oldparagraph*{#1}\mbox{}}
  \newcommand{\xxxParagraphNoStar}[1]{\oldparagraph{#1}\mbox{}}
\fi
\ifx\subparagraph\undefined\else
  \let\oldsubparagraph\subparagraph
  \renewcommand{\subparagraph}{
    \@ifstar
      \xxxSubParagraphStar
      \xxxSubParagraphNoStar
  }
  \newcommand{\xxxSubParagraphStar}[1]{\oldsubparagraph*{#1}\mbox{}}
  \newcommand{\xxxSubParagraphNoStar}[1]{\oldsubparagraph{#1}\mbox{}}
\fi
\makeatother

\usepackage{color}
\usepackage{fancyvrb}

\DefineVerbatimEnvironment{Highlighting}{Verbatim}{commandchars=\\\{\}}
\usepackage{framed}
\definecolor{shadecolor}{RGB}{241,243,245}
\newenvironment{Shaded}{\begin{snugshade}}{\end{snugshade}}

\newcommand{\AttributeTok}[1]{\textcolor[rgb]{0.40,0.45,0.13}{#1}}

\newcommand{\ControlFlowTok}[1]{\textcolor[rgb]{0.00,0.23,0.31}{\textbf{#1}}}

\newcommand{\DecValTok}[1]{\textcolor[rgb]{0.68,0.00,0.00}{#1}}

\newcommand{\FloatTok}[1]{\textcolor[rgb]{0.68,0.00,0.00}{#1}}
\newcommand{\FunctionTok}[1]{\textcolor[rgb]{0.28,0.35,0.67}{#1}}

\newcommand{\NormalTok}[1]{\textcolor[rgb]{0.00,0.23,0.31}{#1}}

\newcommand{\OtherTok}[1]{\textcolor[rgb]{0.00,0.23,0.31}{#1}}

\newcommand{\SpecialCharTok}[1]{\textcolor[rgb]{0.37,0.37,0.37}{#1}}

\newcommand{\StringTok}[1]{\textcolor[rgb]{0.13,0.47,0.30}{#1}}

\providecommand{\tightlist}{%
  \setlength{\itemsep}{0pt}\setlength{\parskip}{0pt}}\usepackage{longtable,booktabs,array}
\usepackage{calc} 
\usepackage{etoolbox}
\makeatletter
\patchcmd\longtable{\par}{\if@noskipsec\mbox{}\fi\par}{}{}
\makeatother
\IfFileExists{footnotehyper.sty}{\usepackage{footnotehyper}}{\usepackage{footnote}}
\makesavenoteenv{longtable}
\usepackage{graphicx}
\makeatletter
\def\maxwidth{\ifdim\Gin@nat@width>\linewidth\linewidth\else\Gin@nat@width\fi}
\def\maxheight{\ifdim\Gin@nat@height>\textheight\textheight\else\Gin@nat@height\fi}
\makeatother
\setkeys{Gin}{width=\maxwidth,height=\maxheight,keepaspectratio}
\makeatletter
\def\fps@figure{htbp}
\makeatother

\usepackage{pmboxdraw}
\makeatletter
\@ifpackageloaded{caption}{}{\usepackage{caption}}
\AtBeginDocument{%
\ifdefined\contentsname
  \renewcommand*\contentsname{Table of contents}
\else
  \newcommand\contentsname{Table of contents}
\fi
\ifdefined\listfigurename
  \renewcommand*\listfigurename{List of Figures}
\else
  \newcommand\listfigurename{List of Figures}
\fi
\ifdefined\listtablename
  \renewcommand*\listtablename{List of Tables}
\else
  \newcommand\listtablename{List of Tables}
\fi
\ifdefined\figurename
  \renewcommand*\figurename{Figure}
\else
  \newcommand\figurename{Figure}
\fi
\ifdefined\tablename
  \renewcommand*\tablename{Table}
\else
  \newcommand\tablename{Table}
\fi
}
\@ifpackageloaded{float}{}{\usepackage{float}}
\floatstyle{ruled}
\@ifundefined{c@chapter}{\newfloat{codelisting}{h}{lop}}{\newfloat{codelisting}{h}{lop}[chapter]}
\floatname{codelisting}{Listing}

\makeatother
\makeatletter
\@ifpackageloaded{caption}{}{\usepackage{caption}}
\@ifpackageloaded{subcaption}{}{\usepackage{subcaption}}
\makeatother

\ifLuaTeX
  \usepackage{selnolig}  
\fi
\usepackage[]{natbib}
\usepackage{bookmark}

\IfFileExists{xurl.sty}{\usepackage{xurl}}{} 
\hypersetup{
  pdftitle={Automated residual plot assessment with the  package  and the  application },
  pdfauthor={Weihao Li; Dianne Cook; Emi Tanaka; Susan VanderPlas; Klaus Ackermann},
  pdfkeywords={initial data analysis, statistical graphics, data
visualisation, visual inference, computer vision, machine
learning, hypothesis testing, regression analysis, model diagnostics},
  colorlinks=true,
  linkcolor={blue},
  filecolor={Maroon},
  citecolor={Blue},
  urlcolor={Blue},
  pdfcreator={LaTeX via pandoc}}

\usepackage{moreverb}
\usepackage{url}
\usepackage{grffile}
\usepackage[UKenglish]{isodate}

\def\volumeyear{2025}

\usepackage{lineno}

\def\firstletters{\bgroup \catcode`-=10 \catcode`(=10 \filA}
\def\filA#1{\filB#1 {\end} }
\def\filB#1#2 {\ifx\end#1\egroup \else#1 \expandafter\filB\fi}

\runningheads{
Automated residual plot assessment
}{
\firstletters{WEIHAO}LI, \firstletters{DIANNE}COOK, \firstletters{EMI}TANAKA, \firstletters{SUSAN}VANDERPLAS, AND  \firstletters{KLAUS}ACKERMANN
}

\title{Automated residual plot assessment with the \textsf{R} package
\textsf{autovi} and the \textsf{Shiny} application \textsf{autovi.web}}

\author{
Weihao Li\addressnum{1, 2},
Dianne Cook\addressnum{1},
Emi Tanaka\addressnum{2},
Susan VanderPlas\addressnum{3} and
Klaus Ackermann\addressnum{1}
}
\affiliation{
Monash University,
The Australian National University and
University of Nebraska
}

\address{
\addressnum{1} Department of Econometrics and Business
Statistics, Monash University, Wellington Road, VIC 3800, Australia\\
\addressnum{2} Research School of Finance, Actuarial Studies and
Statistics, The Australian National University, CBE Building 26C,
Kingsley Street, ACT 2600, Australia\\
\addressnum{3} Department of Statistics, University of Nebraska, Hardin
Hall, 3310 Holdrege St Suite 340, Lincoln, NE 68583, United States\\
\hspace*{1ex} Email: \texttt{patrick.li@anu.edu.au}
}

\date{2025}
\begin{document}

\begin{abstract}
Visual assessment of residual plots is a common approach for diagnosing
linear models, but it relies on manual evaluation, which does not scale
well and can lead to inconsistent decisions across analysts. The lineup
protocol, which embeds the observed plot among null plots, can reduce
subjectivity but requires even more human effort. In today's data-driven
world, such tasks are well-suited for automation. We present a new
\textsf{R} package that uses a computer vision model to automate the
evaluation of residual plots. An accompanying \textsf{Shiny} application
is provided for ease of use. Given a sample of residuals, the model
predicts a visual signal strength (VSS) and offers supporting
information to help analysts assess model fit.
\end{abstract}


\maketitle

\section{Introduction}\label{sec-autovi-introduction}

Regression analysis is a widely used statistical modelling technique for
data in many fields. There is a vast array of software for conducting
regression modelling and generating diagnostics. The package
\textsf{lmtest} \citep{lmtest} provides a suite of conventional tests,
while the \textsf{stats} package \citep{stats} offers standard
diagnostic plots such as residuals vs.~fitted values, quantile-quantile
(Q-Q) plots, and residuals vs.~leverage plots. Additional packages like
\textsf{jtools} \citep{jtools}, \textsf{olsrr} \citep{olsrr},
\textsf{rockchalk} \citep{rockchalk}, and \textsf{ggResidpanel}
\citep{ggresidpanel} deliver similar graphical diagnostics, often with
enhanced aesthetics or interactive features. These tools collectively
produce the core diagnostic plots outlined in the classical text by
\citet{cook1982residuals}. The \textsf{ecostats} package
\citep{warton_global_2023} extends these diagnostics by incorporating
simulation envelopes into residual plots. Meanwhile, \textsf{DHARMa}
\citep{dharma} compares empirical quantiles (0.25, 0.5, and 0.75) of
scaled residuals to their theoretical counterparts, with a strong focus
on identifying model violations such as heteroscedasticity, misspecified
functional forms, and issues specific to generalised linear and
mixed-effect models, like over/under-dispersion. It also provides
conventional test annotations to reduce the risk of misinterpretation.

However, relying solely on subjective assessments of these plots can
lead to issues such as over-interpreting random patterns as model
violations. \citet{li2024plot} demonstrated that visual inference
methods, particularly those using the lineup protocol
\citep{buja2009statistical}, offer more practical and reliable
assessments of residual patterns than conventional tests, as they are
less sensitive to minor departures. Packages such as \textsf{nullabor}
\citep{nullabor}, \textsf{HLMdiag} \citep{loy2014hlmdiag}, and
\textsf{regressinator} \citep{regressinator} support this approach by
enabling users to compare observed residual plots with plots generated
under null hypothesis, thereby helping to quantify the significance of
any detected patterns.

As noted in \citet{li2024automated}, the lineup protocol has significant
limitations in large-scale applications due to its reliance on human
labour. To overcome this constraint, a computer vision model was
developed alongside a corresponding statistical testing procedure to
automate the assessment of residual plots. The model takes as input a
residual plot and a set of auxiliary variables (such as the number of
observations) and outputs a predicted visual signal strength (VSS). This
VSS estimates the degree of deviation between the residual distribution
of the fitted model and the reference distribution expected under
correct model specification.

To make the statistical testing procedure and trained computer vision
model widely accessible, we developed the \textsf{R} package
\textsf{autovi} along with a companion web interface,
\textsf{autovi.web}, which allows users to automatically assess their
residual plots using the trained computer vision model.

The remainder of this paper is structured as follows:
Section~\ref{sec-vss-desc} introduces the definition and computation of
visual signal strength. Section~\ref{sec-null-and-boot-desc} expands on
the computation of the null and bootstrapped residuals.
Section~\ref{sec-autovi} provides a detailed documentation of the
\textsf{autovi} package, including its usage and infrastructure.
Section~\ref{sec-autovi-web} focuses on the \textsf{autovi.web}
interface, describing its design and usage, along with illustrative
examples. Finally, Section~\ref{sec-autovi-conclusion} presents the main
conclusions of this work.

\section{Definition and computation of visual signal
strength}\label{sec-vss-desc}

To train a computer vision model, a measure of the visible pattern in a
plot is needed. We call this the \textbf{visual signal strength} (VSS),
which measures how prominently a specific set of visual patterns appears
in an image. This can be computed for a training set of data, and plots,
where the generating distributions are specified.

In the context of classical normal linear regression model diagnostics,
VSS describes the clarity of visual patterns on a diagnostic plot that
may indicate model violations. Violations can be categorised as weak,
moderate, or strong, but here we treat it as a continuous positive real
variable. Importantly, its interpretation depends on how it is linked to
a function of the data or the underlying data generating process.
Consequently, the calculation of VSS can vary across different model
classes or within the same model, depending on the generating function.

VSS estimates the distance between the residual distribution of a fitted
classical normal linear regression model and a reference distribution
\citep[see][for details]{li2024automated}. The distance measure is based
on the Kullback-Leibler (KL) divergence:

\begin{equation*} \label{eq:kl-0}
D = \log\left(1 + D_{KL}\right),
\end{equation*}

where \(D_{KL}\) is given by:

\begin{equation} \label{eq:kl-1}
D_{KL} = \int_{\mathbb{R}^{n}}\log\frac{p(\hat{\boldsymbol{e}})}{q(\hat{\boldsymbol{e}})}p(\hat{\boldsymbol{e}})d\hat{\boldsymbol{e}},
\end{equation}

here, \(\hat{\boldsymbol{e}}\) denotes the residual vector from the
regression model, and \(p(\cdot)\) and \(q(\cdot)\) are the probability
density functions of the reference residual distribution \(\mathrm{P}\)
and the true residual distribution \(\mathrm{Q}\), respectively.

This distance measure depends on knowledge of the true residual
distribution, which is unknown in practice. To compute \(D_{KL}\) for
the training samples, Equation (\ref{eq:kl-1}) takes different forms
depending on the specific model violations. For instance, where
necessary higher-order predictors, \(\boldsymbol{Z}\), and their
corresponding parameter, \(\boldsymbol{\beta}_Z\), are omitted from the
fitted linear model, the distance measure can be expanded as follows:

\begin{equation*} \label{eq:kl-2}
D_{KL} = \frac{1}{2}\left(\boldsymbol{\mu}_z^\top(\text{diag}(\boldsymbol{R}\sigma^2))^{-1}\boldsymbol{\mu}_z\right),
\end{equation*}

where
\(\boldsymbol{\mu}_z = \boldsymbol{R}\boldsymbol{Z}\boldsymbol{\beta}_z\),
\(\boldsymbol{R} = \boldsymbol{I}_n - \boldsymbol{X}(\boldsymbol{X}^\top\boldsymbol{X})^{-1}\boldsymbol{X}^\top\)
and \(\boldsymbol{X}\) is the design matrix of the regression model.

The computer vision model approximates this mapping from a set of
residuals to its corresponding distance measure. It is trained on a
large number of synthetic regression models, each designed to simulate
specific violations of classical linear regression assumptions. These
models incorporate non-linearity through Hermite polynomial
transformations of predictors, heteroscedasticity by making the error
variance a predictor-dependent function, and non-normality by drawing
residuals from distributions such as discrete, uniform, and lognormal.
Both simple and multiple linear regression structures are used, with
controlled parameters to generate diverse and complex residual patterns.
Since the data-generating process is known, the distance measure \(D\)
can be explicitly calculated, enabling supervised training. The computer
vision model takes a residual plot as input and outputs the
corresponding distance measure, learning to quantify model violations
directly from visual patterns. Additional details are provided in
\citet{li2024automated}.

\section{Definition and simulation of null and bootstrapped
residuals}\label{sec-null-and-boot-desc}

In the subsequent sections, we will frequently refer to null residuals
and bootstrapped residuals, so it is helpful to first define and explain
how they are generated.

\textbf{Null residuals} are used to generate null plots within the
lineup protocol framework, serving as the foundation for the statistical
testing in our automated residual plot assessment. Specifically, they
represent residuals generated under the null hypothesis that the model
is correctly specified. A common method for simulating null residuals in
linear regression involves sampling from a normal distribution with mean
zero and variance equal to the estimated variance of the error term.
These simulated residuals and their corresponding plots depict what one
would expect from a correctly specified model. If the true residual plot
exhibits noticeable deviations from these null plots, it may suggest
model misspecification.

Our computer vision model is trained to assign lower VSS to null plots
and higher VSS to plots that display distinct patterns. Accordingly,
statistical testing is performed by computing the proportion of null
plots whose VSS equals or exceeds that of the observed residual plot.
This proportion serves as a p-value for a one-sided hypothesis test.

\textbf{Bootstrapped residuals} are obtained by refitting the model on
bootstrap samples, which are generated by sampling individual
observations with replacement from the original dataset. The residual
plots obtained from these refitted models are evaluated using the same
computer vision model. The predicted VSS from the bootstrapped plots
provide an empirical estimate of the variation in the VSS of the
observed residual plot. By examining the proportion of bootstrapped
plots that also exhibit significant violations, we can assess whether
the original conclusion is robust to sampling variability.

\section{\texorpdfstring{\textsf{R} package:
\textsf{autovi}}{ package: }}\label{sec-autovi}

The main purpose of \textsf{autovi} is to provide rejection decisions
and \(p\)-values for testing the null hypothesis (\(H_0\)) that the
regression model is correctly specified. The package provides automated
interpretation of residual plots using computer vision. The name
\textsf{autovi} stands for \textbf{auto}mated \textbf{v}isual
\textbf{i}nference. This functionality can be accessed through the
\textsf{R} package \textsf{autovi}, or through a web interface,
\textsf{autovi.web}, which enables users to perform analyses without
installing \textsf{R}, \textsf{Python}, or their associated dependencies
locally.

\subsection{Motivation}\label{sec-why}

\begin{figure}

\centering{

\includegraphics[width=1\textwidth,height=\textheight]{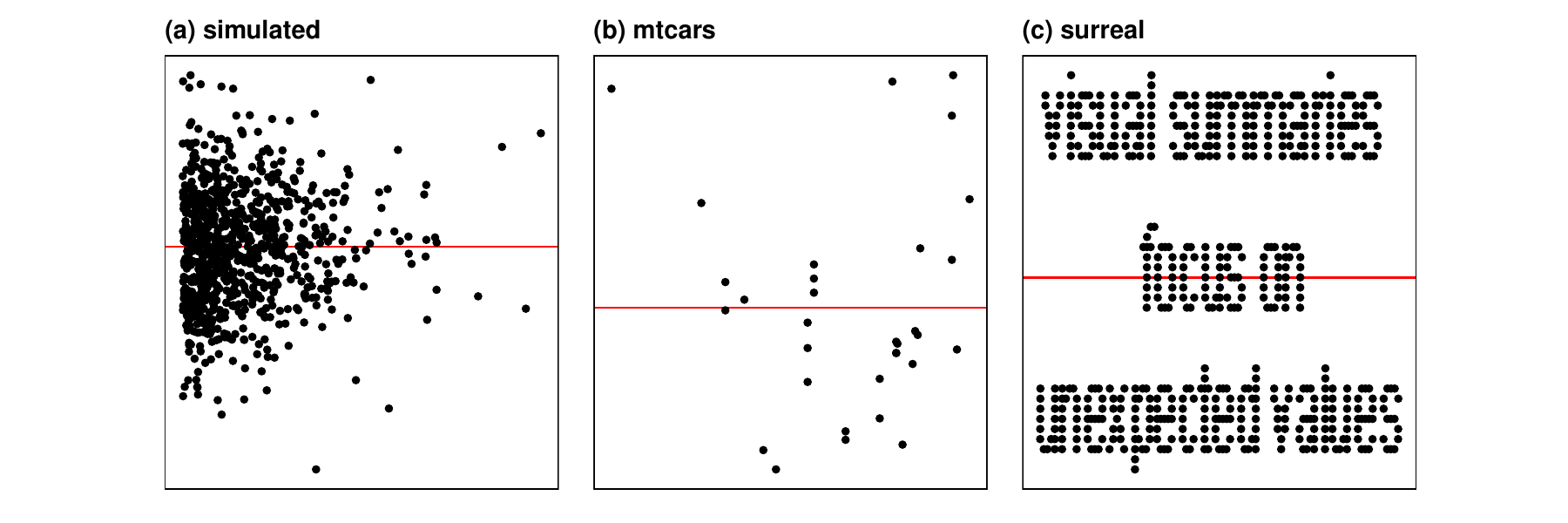}

}

\caption{\label{fig-three-examples}Reading residual plots can be a
difficult task, particularly for students new to statistical modelling.
The \textsf{autovi} package makes it easier. Here are three examples of
residual plots, which may appear to have structure. According to autovi,
the visual signal strengths (VSS) of these three examples are
approximately (a) 1.53, (b) 3.57, (c) 5.87, resulting in (b), (c) being
significant violations of good residuals, but (a) is consistent with a
good residual plot.}

\end{figure}

Figure~\ref{fig-three-examples} shows three sets of plots of residuals
against fitted values. The simulated example in (a) might be interpreted
as a heteroscedastic pattern, however the automated reading would
predict this to have a visual signal strength (VSS) of 1.53, with a
corresponding \(p\)-value of 0.25. This means it would be interpreted as
a good residual plot, that there is nothing in the data to indicate a
violation of model assumptions. Skewness in the predictor variables is
generating the apparent heteroscedasticity, where the smaller variance
in residuals at larger fitted values is due to smaller sample size only.
The Breusch--Pagan test \citep{breusch1979simple} for heteroscedasticity
would also not reject this as good residual plot.

The data in (b) is generated by fitting a linear model predicting
\texttt{mpg} based on \texttt{hp} using the
\textsf{datasets}\texttt{::mtcars}. It is a small data set, and there is
a hint of nonlinear structure not captured by the model. The automated
plot reading would predict a VSS of 3.57, which has a \(p\)-value less
than 0.05. That is, the nonlinear structure is most likely real, and
indicates a problem with the model. The conventional test, a Ramsey
Regression Equation Specification Error Test (RESET)
\citep{ramsey1969tests} would also strongly detect the nonlinearity.

The third example is generated using the \textsf{surreal} package
\citep{surreal}, where structure residuals are embedded in the data. In
this case, a quote inspired by Tukey, `visual summaries focus on
unexpected values', is used to define the residual structure. The
automated plot reading predicts the VSS to be 5.87, with a \(p\)-value
less than 0.05. Visually, the structure is strikingly clear, but a RESET
test for nonlinear structure would not report a problem. (It would be
detected by a Breusch--Pagan for heteroscedasticity and also
Shapiro--Wilk test \citep{shapiro1965analysis} for non-normality.)

\subsection{Implementation}\label{sec-autovi-implementation}

The \textsf{autovi} package is built on the \textsf{bandicoot}
object-oriented programming (OOP) system \citep{bandicoot}, marking a
departure from \textsf{R}'s traditional S3 generic system. This OOP
architecture enhances flexibility and modularity, allowing users to
redefine key functions through method overriding.

The \textsf{autovi} infrastructure effectively integrates multiple
programming languages and libraries into a comprehensive analytical
tool. It relies on five core libraries from \textsf{Python} and
\textsf{R}, each playing a critical role in the analysis pipeline. In
\textsf{Python}, \textsf{pillow} \citep{clark2015pillow} handles image
processing tasks such as reading and resizing PNG files of residual
plots, then converting them into input tensors for further analysis.
\textsf{TensorFlow} \citep{abadi2016tensorflow}, a key component of
modern machine learning, is used to predict the VSS of these plots using
a pre-trained convolutional neural network.

In the \textsf{R} environment, \textsf{autovi} utilizes several
libraries. \textsf{ggplot2} \citep{ggplot2} generates the initial
residual plots, saved as PNG files for visual input. \textsf{cassowaryr}
\citep{mason2022cassowaryr} computes scagnostics (scatter plot
diagnostics), providing numerical features that capture statistical
properties of the plots. These scagnostics complement the visual
analysis by offering quantitative metrics as secondary input to the
computer vision model. \textsf{reticulate} \citep{reticulate} enables
seamless communication between \textsf{R} and \textsf{Python}.

\subsection{Installation}\label{installation}

The \textsf{autovi} package is available on CRAN. It is actively
developed and maintained, with the latest updates accessible on GitHub.
This paper uses \textsf{autovi} version 0.4.2. The package includes
internal functions to check the current \textsf{Python} environment used
by the \textsf{reticulate} package. If the necessary \textsf{Python}
packages are not installed in the \textsf{Python} interpreter, an error
will be raised. If you want to select a specific \textsf{Python}
environment, you can do so by calling the
\textsf{reticulate}\texttt{::use\_python()} function before using the
\textsf{autovi} package.

We recommend using the \textsf{Shiny} application \textsf{autovi.web} if
users encounter installation problems.

\subsection{Usage}\label{sec-autovi-usage}

\subsubsection{Numerical summary}\label{sec-autovi-numerical}

Three steps are needed to get an automated assessment of a set of
residuals and fitted values:

\begin{enumerate}
\def\labelenumi{\arabic{enumi}.}
\tightlist
\item
  Load the \textsf{autovi} package using the \texttt{library()}
  function.
\item
  Create a checker object with a linear regression model.
\item
  Call the \texttt{check()} method of the checker, which, by default,
  predicts the VSS for the true residual plot, 100 null plots, and 100
  bootstrapped plots. The method stores the predictions internally and
  prints a concise results report.
\end{enumerate}

The code to do this is:

\begin{Shaded}
\begin{Highlighting}[]
\FunctionTok{library}\NormalTok{(autovi) }
\NormalTok{checker }\OtherTok{\textless{}{-}} \FunctionTok{residual\_checker}\NormalTok{(}\FunctionTok{lm}\NormalTok{(dist }\SpecialCharTok{\textasciitilde{}}\NormalTok{ speed, }\AttributeTok{data =}\NormalTok{ cars))}
\NormalTok{checker}\SpecialCharTok{$}\FunctionTok{check}\NormalTok{() }
\end{Highlighting}
\end{Shaded}

It produces the following summary:

\begin{verbatim}
<AUTO_VI object>
 Status:
  - Fitted model: lm
  - Keras model: (None, 32, 32, 3) + (None, 5) -> (None, 1)
     - Output node index: 1
  - Result:
     - Observed visual signal strength: 3.16 (p-value = 0.0396)
     - Null visual signal strength: [100 draws]
        - Mean: 1.274
        - Quantiles:
           ╔══════════════════════════════════════════╗
           ║  25%   50%   75%   80%   90%   95%   99% ║
           ║0.802 1.111 1.575 1.666 1.919 2.657 3.348 ║
           ╚══════════════════════════════════════════╝
     - Bootstrapped visual signal strength: [100 draws]
        - Mean: 2.795 (p-value = 0.05941)
        - Quantiles:
           ╔══════════════════════════════════════════╗
           ║  25%   50%   75%   80%   90%   95%   99% ║
           ║2.455 2.941 3.177 3.300 3.474 3.537 3.668 ║
           ╚══════════════════════════════════════════╝
     - Likelihood ratio: 0.7333 (boot) / 0.06284 (null) = 11.67
\end{verbatim}

The summary includes observed VSS of the true residual plot and
associated \(p\)-value of the automated visual test. The \(p\)-value is
the proportion of null plots (out of the total 100) that have VSS
greater than or equal to that of the true residual plot. The report also
provides sample quantiles of VSS for null samples and bootstrapped data
plots, providing more information about the sampling variability and a
likelihood of model violations. The likelihood is computed from the
proportion of values greater than the observed VSS in both the
bootstrapped data values and the simulated null values.

\subsubsection{Visual summary}\label{sec-autovi-visual}

Users can visually inspect the original residual plot alongside a sample
null plot using \texttt{plot\_pair()} or a lineup of null plot
\texttt{plot\_lineup()}. This visual comparison can clarify why \(H_0\)
is either rejected or not, and help identify potential remedies.

\begin{Shaded}
\begin{Highlighting}[]
\NormalTok{checker}\SpecialCharTok{$}\FunctionTok{plot\_pair}\NormalTok{()}
\end{Highlighting}
\end{Shaded}

\begin{figure}[H]

\centering{

\includegraphics[width=0.45\textwidth,height=\textheight]{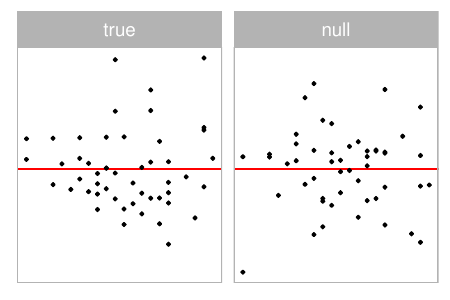}

}

\caption{\label{fig-plot-pair}True plot alongside one null plot, for
quick comparison.}

\end{figure}%

The \texttt{plot\_pair()} method (Figure~\ref{fig-plot-pair}) displays
the true residual plot on the left and a single null plot on the right.
If a full lineup was shown, the true residual plot would be embedded in
a page of null plots. Users should look for any distinct visual patterns
in the true residual plot that are absent in the null plot. Running
these functions multiple times can help any visual suspicions, as each
execution generates new random null plots for comparison.

The package offers a straightforward visualisation of the assessment
result through the \texttt{summary\_plot()} function.

\begin{Shaded}
\begin{Highlighting}[]
\NormalTok{checker}\SpecialCharTok{$}\FunctionTok{summary\_plot}\NormalTok{()}
\end{Highlighting}
\end{Shaded}

\begin{figure}[H]

\centering{

\includegraphics[width=1\textwidth,height=\textheight]{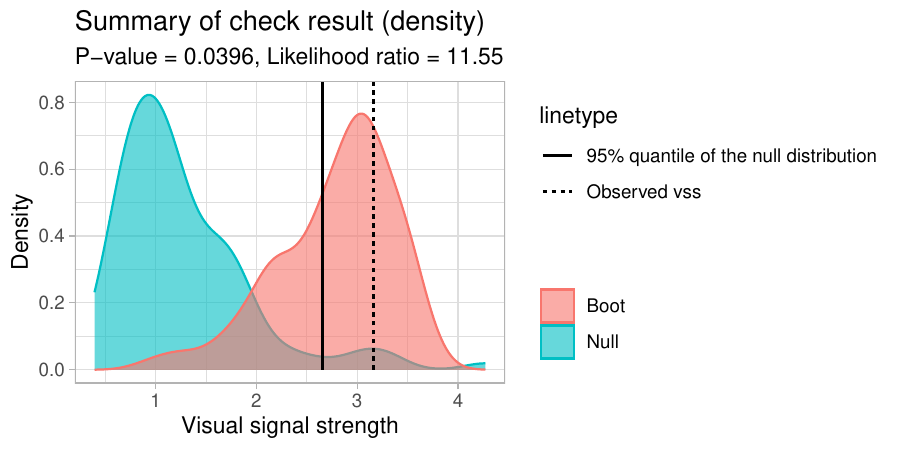}

}

\caption{\label{fig-summary-plot}Summary plot comparing the densities of
VSS for bootstrapped residual samples (red) relative to VSS for null
plots (blue).}

\end{figure}%

In the result, shown in Figure~\ref{fig-summary-plot}, the blue area
represents the density of VSS for null residual plots, while the red
area shows the density for bootstrapped residual plots. The dashed line
indicates the VSS of the true residual plot, and the solid line marks
the critical value at a 95\% significance level. The \(p\)-value and the
likelihood ratio are displayed in the subtitle. The likelihood ratio
represents the ratio of the likelihood of observing the VSS of the true
residual plot from the bootstrapped distribution compared to the null
distribution.

Interpreting the plot involves several key aspects. If the dashed line
falls to the right of the solid line, it suggests rejecting the null
hypothesis. The degree of overlap between the red and blue areas
indicates similarity between the true residual plot and null plots;
greater overlap suggests more similarity. Lastly, the portion of the red
area to the right of the solid line represents the percentage of
bootstrapped models considered to have model violations.

This visual summary provides an intuitive way to assess the model's fit
and potential violations, allowing users to quickly grasp the results of
the automated analysis.

\subsection{Modularized infrastructure}\label{sec-autovi-infrastructure}

\begin{figure}

\centering{

\includegraphics[width=1\textwidth,height=\textheight]{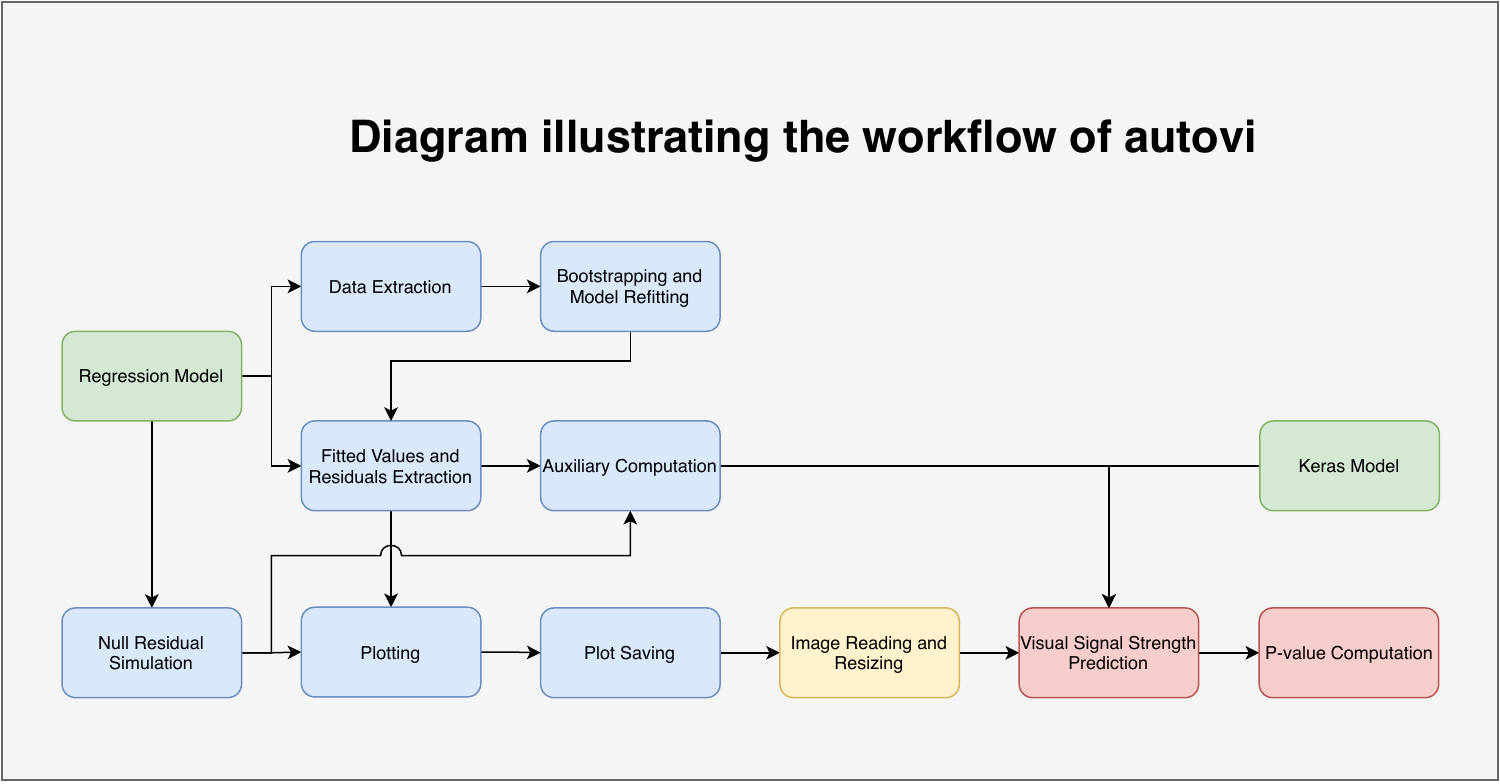}

}

\caption{\label{fig-autovi-diag}Diagram illustrating the infrastructure
of the \textsf{R} package \textsf{autovi}. The modules in green are
primary inputs provided by users. Modules in blue are overridable
methods that can be modified to accommodate users' specific needs. The
module in yellow is a pre-defined non-overridable method. The modules in
red are primary outputs of the package.}

\end{figure}%

The initial motivation for developing \textsf{autovi} was to create a
convenient interface for sharing the models described and trained in
\citet{li2024automated}. However, recognising that the classical normal
linear regression model represents a restricted class of models, we
sought to avoid limiting the potential for future extensions, whether by
the original developers or other developers. As a result, the package
was designed to function seamlessly with linear regression models with
minimal modification and few required arguments, while also
accommodating other classes of models through partial infrastructure
substitution. This modular and customisable design allows
\textsf{autovi} to handle a wide range of residual diagnostics tasks.

The infrastructure of \textsf{autovi} consists of ten core modules: data
extraction, bootstrapping and model refitting, fitted values and
residuals extraction, auxiliary computation, null residual simulation,
plotting, plot saving, image reading and resizing, VSS prediction, and
\(p\)-value computation. Each module is designed with minimal dependency
on the preceding modules, allowing users to customise parts of the
infrastructure without affecting its overall integrity. An overview of
this infrastructure is illustrated in Figure~\ref{fig-autovi-diag}.

The package takes regression models and a \textsf{Keras} model as
primary inputs. Modules for VSS prediction and \(p\)-value computation
are fixed but accessible via function arguments, using
\textsf{TensorFlow} for inference and statistical testing. The image
loading module is also fixed, using \textsf{PIL} to read and resize
images based on the \textsf{Keras} model's input shape. The remaining
seven modules are overridable, allowing users to adapt the workflow as
needed. The data extraction module extracts a \texttt{data.frame}
containing variables used in the regression model. The bootstrapping and
refitting module resamples the data and refits the model. The fitted
values and residuals extraction module returns these values as a
\texttt{data.frame}. The auxiliary computation module calculates
scagnostics such as monotonicity. The plotting module generates a
\texttt{ggplot} in a standard format, and the plot saving module exports
it at the same resolution as the training images. These modules are
described in detail in the package documentation.

\subsection{Extension to other model
classes}\label{extension-to-other-model-classes}

The \textsf{autovi} \textsf{R} package can be extended to accommodate
other classes of models beyond linear regression, such as generalised
linear models (\texttt{glm}). This is achieved by substituting the
relevant overridable modules, and if needed, supplying a different
\textsf{Keras} model.

We provide an example of defining a new checker class tailored for
Poisson regression using the \texttt{glm} framework:

\begin{enumerate}
\def\labelenumi{\arabic{enumi}.}
\tightlist
\item
  Define a new class using \texttt{new\_class()} with \texttt{AUTO\_VI}
  as the parent class.
\item
  Override the necessary methods using \texttt{register\_method()}. In
  this example, we use Pearson residuals. To simulate null residuals, we
  assume the fitted model is correct and the estimated coefficients are
  accurate. New response values are generated accordingly, and a new
  model is fitted to this simulated response. Null residuals are then
  extracted from this refitted model.
\item
  Create an alias for the \texttt{instantiate()} method of the new
  class.
\end{enumerate}

\begin{Shaded}
\begin{Highlighting}[]
\NormalTok{AUTO\_POIS\_VI }\OtherTok{\textless{}{-}} \FunctionTok{new\_class}\NormalTok{(AUTO\_VI, }\AttributeTok{class\_name =} \StringTok{"AUTO\_POIS\_VI"}\NormalTok{)}
\FunctionTok{register\_method}\NormalTok{(}
\NormalTok{  AUTO\_POIS\_VI,}
  \AttributeTok{get\_fitted\_and\_resid =} \ControlFlowTok{function}\NormalTok{(}\AttributeTok{fitted\_model =}\NormalTok{ self}\SpecialCharTok{$}\NormalTok{fitted\_model) \{}
    \FunctionTok{tibble}\NormalTok{(}\AttributeTok{.fitted =} \FunctionTok{fitted}\NormalTok{(fitted\_model),}
           \AttributeTok{.resid =} \FunctionTok{resid}\NormalTok{(fitted\_model, }\AttributeTok{type =} \StringTok{"pearson"}\NormalTok{))}
\NormalTok{  \},}
  \AttributeTok{null\_method =} \ControlFlowTok{function}\NormalTok{(}\AttributeTok{fitted\_model =}\NormalTok{ self}\SpecialCharTok{$}\NormalTok{fitted\_model) \{}
\NormalTok{    dat }\OtherTok{\textless{}{-}} \FunctionTok{model.frame}\NormalTok{(fitted\_model)}
\NormalTok{    dat[[}\DecValTok{1}\NormalTok{]] }\OtherTok{\textless{}{-}} \FunctionTok{rpois}\NormalTok{(}\FunctionTok{nrow}\NormalTok{(dat), }\AttributeTok{lambda =} \FunctionTok{fitted}\NormalTok{(fitted\_model))}
\NormalTok{    new\_mod }\OtherTok{\textless{}{-}} \FunctionTok{update}\NormalTok{(fitted\_model, }\AttributeTok{data =}\NormalTok{ dat)}
    \FunctionTok{return}\NormalTok{(self}\SpecialCharTok{$}\FunctionTok{get\_fitted\_and\_resid}\NormalTok{(new\_mod))}
\NormalTok{  \}}
\NormalTok{)}
\NormalTok{auto\_pois\_vi }\OtherTok{\textless{}{-}}\NormalTok{ AUTO\_POIS\_VI}\SpecialCharTok{$}\NormalTok{instantiate}
\end{Highlighting}
\end{Shaded}

\begin{figure}[H]

\centering{

\includegraphics[width=0.8\textwidth,height=\textheight]{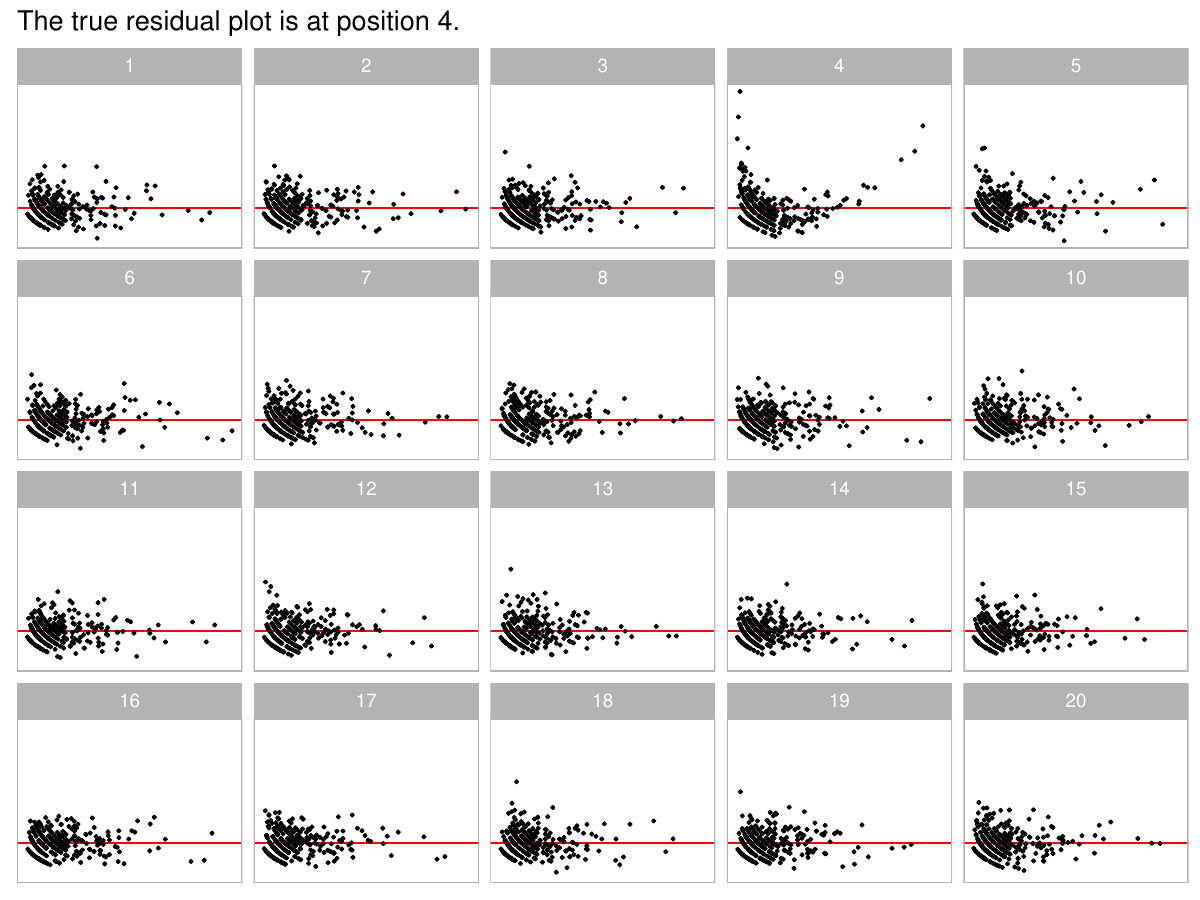}

}

\caption{\label{fig-pois-lineup}A lineup of residual plots from Poisson
generalised linear models, with the true residual plot at position 4,
which displays a distinct U-shaped pattern. In contrast, the null plots
show characteristics broadly consistent with well-behaved residuals from
linear regression models.}

\end{figure}

The resulting checker class can be employed analogously to the linear
model case described in Section~\ref{sec-autovi-usage}. For
illustration, we fit a Poisson model in which the quadratic term of the
predictor \(x\) is intentionally omitted. This misspecification
manifests as a pronounced U-shaped pattern in the lineup display (see
Figure~\ref{fig-pois-lineup}), which is also successfully identified by
the computer vision model, yielding a p-value substantially below the
conventional threshold of 0.05.

\begin{Shaded}
\begin{Highlighting}[]
\NormalTok{x }\OtherTok{\textless{}{-}} \FunctionTok{rnorm}\NormalTok{(}\DecValTok{300}\NormalTok{, }\AttributeTok{sd =} \FloatTok{0.5}\NormalTok{)}
\NormalTok{y }\OtherTok{\textless{}{-}} \FunctionTok{rpois}\NormalTok{(}\DecValTok{300}\NormalTok{, }\AttributeTok{lambda =} \FunctionTok{exp}\NormalTok{(}\DecValTok{1} \SpecialCharTok{+}\NormalTok{ x }\SpecialCharTok{+}\NormalTok{ x}\SpecialCharTok{\^{}}\DecValTok{2}\NormalTok{))}
\NormalTok{pois\_checker }\OtherTok{\textless{}{-}} \FunctionTok{auto\_pois\_vi}\NormalTok{(}
  \FunctionTok{glm}\NormalTok{(y }\SpecialCharTok{\textasciitilde{}}\NormalTok{ x, }\AttributeTok{family =} \StringTok{"poisson"}\NormalTok{),}
  \AttributeTok{keras\_model =} \FunctionTok{get\_keras\_model}\NormalTok{(}\StringTok{"vss\_phn\_32"}\NormalTok{)}
\NormalTok{)}
\NormalTok{pois\_checker}\SpecialCharTok{$}\FunctionTok{plot\_lineup}\NormalTok{()}
\end{Highlighting}
\end{Shaded}

\begin{Shaded}
\begin{Highlighting}[]
\NormalTok{pois\_checker}\SpecialCharTok{$}\FunctionTok{check}\NormalTok{()}
\end{Highlighting}
\end{Shaded}

\begin{verbatim}
<AUTO_POIS_VI object>
 Status:
  - Fitted model: glm, lm
  - Keras model: (None, 32, 32, 3) + (None, 5) -> (None, 1)
     - Output node index: 1
  - Result:
     - Observed visual signal strength: 4.875 (p-value = 0.009901)
     - Null visual signal strength: [100 draws]
        - Mean: 1.331
        - Quantiles:
           ╔══════════════════════════════════════════╗
           ║  25%   50%   75%   80%   90%   95%   99% ║
           ║1.035 1.233 1.488 1.644 1.941 2.276 2.639 ║
           ╚══════════════════════════════════════════╝
     - Bootstrapped visual signal strength: [100 draws]
        - Mean: 5.51 (p-value = 0.009901)
        - Quantiles:
           ╔══════════════════════════════════════════╗
           ║  25%   50%   75%   80%   90%   95%   99% ║
           ║5.330 5.505 5.698 5.735 5.830 5.903 6.013 ║
           ╚══════════════════════════════════════════╝
     - Likelihood ratio: 0.05096 (boot) / 0 (null) = Extremely large
\end{verbatim}

It is important to note, however, that the pre-trained computer vision
model included in \textsf{autovi}, such as \texttt{vss\_phn\_32} (see
\texttt{list\_keras\_models()} for the full list of available models),
was developed specifically for diagnostics of linear regression. Its
applicability to other model classes relies on the assumption that the
null residual plots exhibit characteristics broadly consistent with
those of well-behaved linear regression residuals, that is, residuals
should be approximately randomly scattered around zero, display roughly
constant variance across the range of fitted values, and exhibit no
discernible structure or curvature. If these conditions are not met, or
if model violations do not give rise to visually detectable patterns,
the validity of the automated diagnostics may be compromised. In such
cases, users are encouraged to train and apply their own \textsf{Keras}
models. Detailed guidance on model training and discussion on extending
the methodology to other model classes can be found in
\citet{li2024automated}.

\section{\texorpdfstring{Web interface:
\textsf{autovi.web}}{Web interface: }}\label{sec-autovi-web}

The \textsf{autovi.web} \textsf{Shiny} application extends the
functionality of \textsf{autovi} by offering a user-friendly web
interface for automated residual plot assessment. This eliminates the
common challenges associated with software installation, so users can
avoid managing \textsf{Python} environments or handling version
requirements for \textsf{R} libraries. The platform is cross-platform
and accessible on various devices and operating systems, making it
suitable even for users without \textsf{R} programming experience.
Additionally, updates are managed centrally, ensuring that users always
have access to the latest features. This section discusses the
implementation based on \textsf{autovi.web} version 0.1.0.

\subsection{Implementation}\label{implementation}

The interface \textsf{autovi.web} is built using the \textsf{shiny}
\citep{shiny} and \textsf{shinydashboard} \citep{shinydashboard}
\textsf{R} packages. Hosted on the
\href{https://www.shinyapps.io}{shinyapps.io} domain, the application is
accessible through any modern web browser. The \textsf{R} packages
\textsf{htmltools} \citep{htmltools} and \textsf{shinycssloaders}
\citep{shinycssloaders} are used to render markdown documentation in
\textsf{Shiny} application, and for loading animations for
\textsf{Shiny} widgets, respectively.

Determining the best way to implement the backend was difficult. In our
initial planning for \textsf{autovi.web}, we considered implementing the
entire web application using the \textsf{webr} framework \citep{webr},
which would have allowed the entire application to run directly in the
user's browser. However, \textsf{webr} does not support packages which
use compiled \textsf{Fortran} code, which is required by
\textsf{splancs} \citep{splancs}, a dependency of \textsf{autovi}. In
the future, it is possible that a working \textsf{Emscripten}
\citep{zakai2011emscripten} version of this package may allow full
\textsf{webr} support.

We also explored the possibility of implementing the web interface using
frameworks built on other languages, such as \textsf{Python}. However,
server hosting domains that natively support \textsf{Python} servers
typically do not have the latest version of \textsf{R} installed.
Additionally, calling \textsf{R} from \textsf{Python} is typically done
using the \textsf{rpy2} \textsf{Python} library \citep{rpy2}, but this
approach can be awkward when dealing with language syntax related to
non-standard evaluation. Another option we considered was renting a
server where we could have full control, such as those provided by cloud
platforms like Google Cloud Platform (GCP) or Amazon Web Services (AWS).
However, deploying and maintaining the server securely requires some
expertise. Ultimately, the most practical solution was to use the
\textsf{shiny} and \textsf{shinydashboard} frameworks, which are
well-established in the \textsf{R} community and offer a solid
foundation for web application development.

The server-side configuration of \textsf{autovi.web} is carefully
designed to support its functionality. Most required \textsf{Python}
libraries, including \textsf{pillow} and \textsf{numpy}, are
pre-installed on the server. These libraries are integrated into the
\textsf{Shiny} application using the \textsf{reticulate} package, which
provides an interface between \textsf{R} and \textsf{Python}.

Due to \href{https://www.shinyapps.io}{shinyapps.io}'s resource policy,
inactive servers enter sleep mode, clearing the local \textsf{Python}
environment. When reactivated for a new session, libraries must be
reinstalled. While this ensures a clean environment for each session, it
may lead to slightly longer loading times for the first user after a
period of inactivity.

In contrast to \textsf{autovi}, \textsf{autovi.web} leverages
\textsf{TensorFlow.js}, a \textsf{JavaScript} library that allows the
execution of machine learning models directly in the browser. This
choice enables native browser execution, enhancing compatibility across
different user environments, and shifts the computational load from the
server to the client-side. \textsf{TensorFlow.js} also offers better
scalability and performance, especially when dealing with
resource-intensive computer vision models on the web.

While \textsf{autovi} requires downloading the pre-trained computer
vision models from GitHub, these models in `.keras' file format are
incompatible with \textsf{TensorFlow.js}. Therefore, we extract and
store the model weights in JSON files and include them as extra
resources in the \textsf{Shiny} application. When the application
initialises, \textsf{TensorFlow.js} rebuilds the computer vision model
using these pre-stored weights.

To allow communication between \textsf{TensorFlow.js} and other
components of the \textsf{Shiny} application, the \textsf{shinyjs}
\textsf{R} package \citep{shinyjs} is used. This package allows calling
custom \textsf{JavaScript} code within the \textsf{Shiny} framework. The
specialised \textsf{JavaScript} code for initialising
\textsf{TensorFlow.js} and calling \textsf{TensorFlow.js} for VSS
prediction is deployed alongside the \textsf{Shiny} application as
additional resources.

\subsection{Usage}\label{sec-autovi-web-workflow}

The workflow of \textsf{autovi.web} is designed to be straightforward,
with numbered steps displayed in each panel. There are two example
datasets provided by the web application. The single residual plot
example uses the \texttt{dino} dataset from the \textsf{R} package
\textsf{datasauRus} \citep{datasaurus}. The lineup example uses
residuals from a simulated regression model that has a non-linearity
issue. We walk through the lineup example to further demonstrate the
workflow of the web application.

\subsubsection{Reading data and setting
parameters}\label{reading-data-and-setting-parameters}

\begin{figure}

\centering{

\includegraphics[width=1\textwidth,height=\textheight]{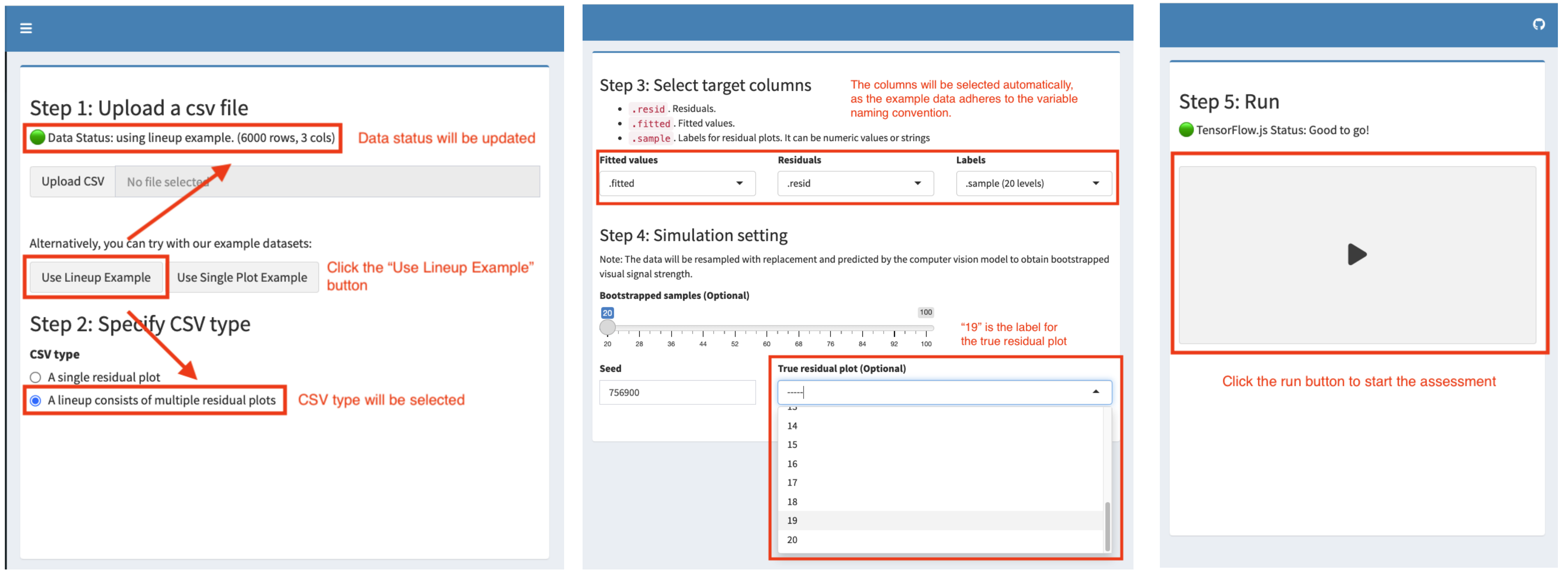}

}

\caption{\label{fig-autovi-web-setup}To begin the workflow for
\textsf{autovi} using the lineup example dataset, the user clicks the
`Use Lineup Example' button (left) to load the example dataset, during
which the data status and CSV type will be automatically updated. The
user must manually select the label for the true residual plot (middle)
to compute further results. The user initiates the assessment of the
lineup example data by clicking the run button (right).}

\end{figure}

The user can select to upload data as either a single set of residuals
and fitted values in a two (or more) column CSV file or a pre-computed
lineup of residuals and null datasets in a three (or more) column CSV
file (i.e.~multiple sets of residuals and fitted values with a column
indicating the set label). Here we illustrate use with lineup example
data sets (Figure~\ref{fig-autovi-web-setup}). To use the lineup example
data, click the `Use Lineup Example' button. The data status will then
update to show the number of rows and columns in the dataset, and the
CSV type will automatically be selected to the correct option. Since the
example dataset follows the variable naming conventions assumed by the
web application, the columns for fitted values, residuals, and labels of
residual plots are automatically mapped such that the column named as
\texttt{.fitted} is mapped to fitted values, \texttt{.resid} is mapped
to residuals and if applicable, \texttt{.sample} to labels of the
residual set (middle image). If the user is working with a custom
dataset, these options must be set accordingly. Whenever a data
containing a lineup, the user must manually select the label for the
true residual plot, otherwise the web application does not provide all
the results. The last step is to click the play button (right image) to
start the assessment.

\subsubsection{Results provided}\label{results-provided}

\begin{figure}

\centering{

\includegraphics[width=1\textwidth,height=\textheight]{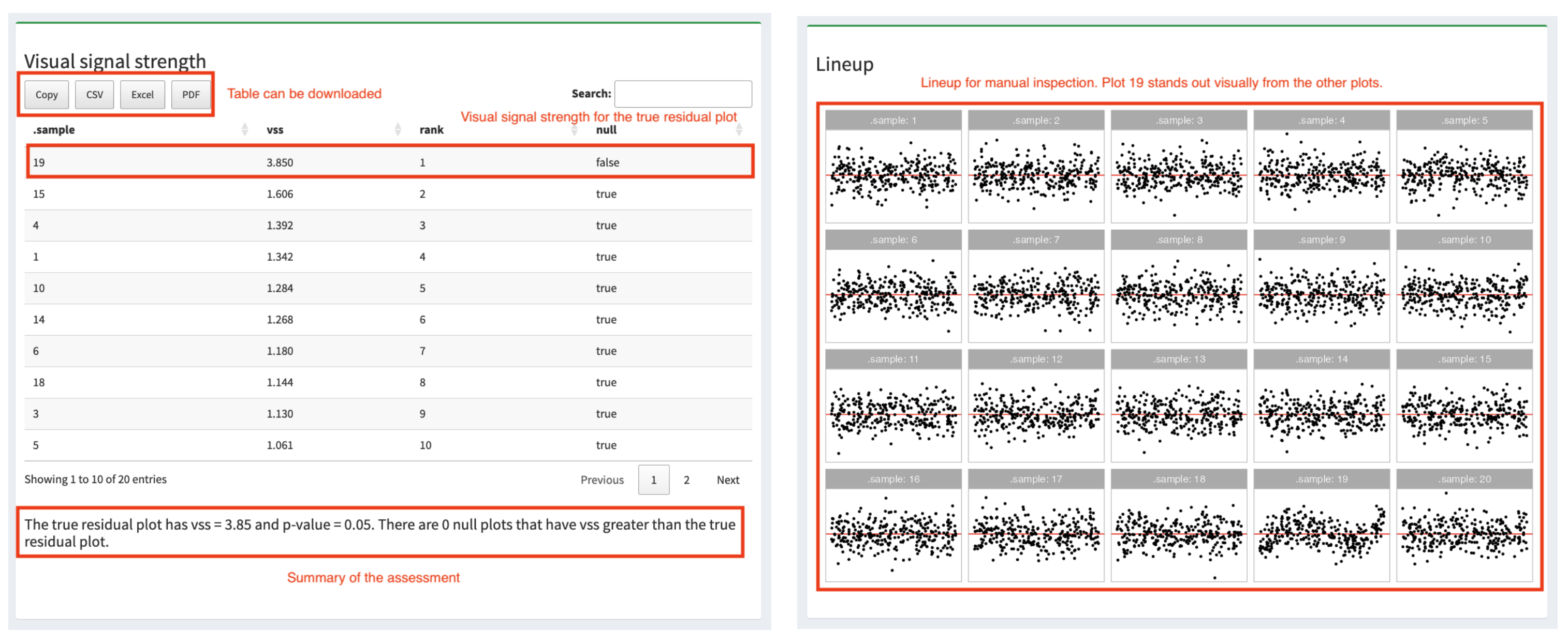}

}

\caption{\label{fig-autovi-web-lineup}Results for the lineup. The VSS of
the true residual plot is displayed in the first row of the table of VSS
values for all the null plots (left image), with a summary text beneath
the table providing the \(p\)-value to aid in decision-making. A lineup
of residual plots allows for manual inspection (right image).}

\end{figure}

Results are provided in multiple panels. The first row of the table
Figure~\ref{fig-autovi-web-lineup} is the most crucial to check, as it
provides the VSS and the rank of the true residual plot among the other
plots. The summary text beneath the table provides the \(p\)-value,
which can be used for quick decision-making. The lineup is for manual
inspection, and the user should see if the true residual plot is
visually distinguishable from the other plots, to confirm if the model
violation is serious.

\begin{figure}

\centering{

\includegraphics[width=1\textwidth,height=\textheight]{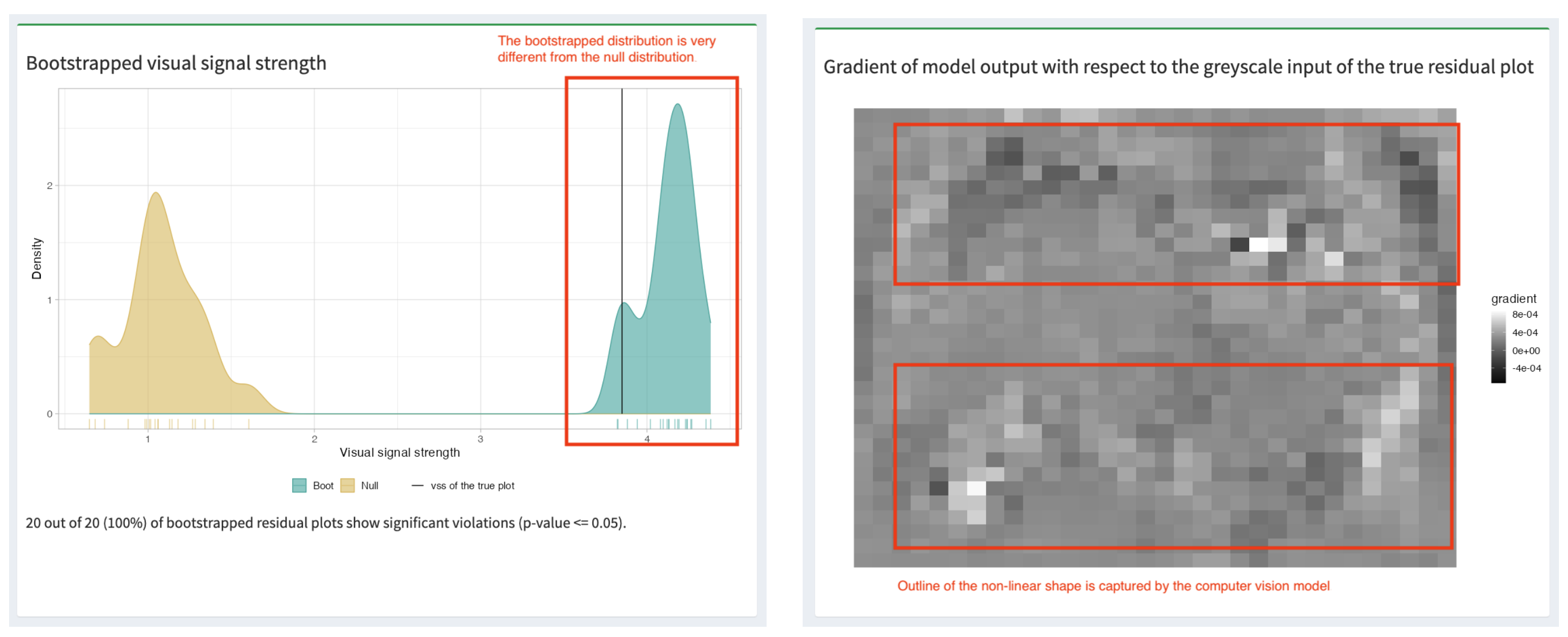}

}

\caption{\label{fig-autovi-web-distributions}Summaries assessing the
strength of the pattern and which elements of the plot contribute. The
density plot helps verify if the bootstrapped distribution differs from
the null distribution (left image). The attention map (right image)
offers insights into whether the computer vision model has captured the
intended visual features of the true residual plot.}

\end{figure}

The density plot in Figure~\ref{fig-autovi-web-distributions} offers a
more robust result, allowing the user to compare the distribution of
bootstrapped VSS with the distribution of null VSS. Finally, the
grayscale attention map (right image) can be used to check if the target
visual features, like the non-linearity present in the lineup example,
are captured by the computer vision model, ensuring the quality of the
assessment. The attention map is the gradient of the model output with
respect to the grayscale image input, indicating the sensitivity of the
output to each pixel.

\section{Conclusions}\label{sec-autovi-conclusion}

This paper presents new regression diagnostics software, the \textsf{R}
package \textsf{autovi} and its accompanying web interface,
\textsf{autovi.web}. It addresses a critical gap in the current
landscape of statistical software. While regression tools are widely
available, effective and efficient diagnostic methods have lagged
behind, particularly in the field of residual plot interpretation.

The \textsf{autovi} \textsf{R} package, introduced in this paper,
automates the assessment of residual plots by incorporating a computer
vision model, reducing reliance on time-consuming and potentially
inconsistent human interpretation. This automation improves the
efficiency of the diagnostic process and promotes consistency in model
evaluation across different users and studies.

The development of the accompanying \textsf{Shiny} app,
\textsf{autovi.web}, expands access to these advanced diagnostic tools,
by providing a user-friendly interface. It makes automated residual plot
assessment accessible to a broader audience, including those who may not
have extensive programming experience. This web-based solution
effectively addresses the potential barriers to adoption, such as
complex dependencies and installation requirements, that are often
associated with advanced statistical software.

The combination of \textsf{autovi} and \textsf{autovi.web} offers a
comprehensive solution to the challenges of residual plot interpretation
in regression analysis. These tools have the potential to significantly
improve the quality and consistency of model diagnostics across various
fields, from academic research to industry applications. By automating a
critical aspect of model evaluation, they allow researchers and analysts
to focus more on interpreting results and refining models, rather than
grappling with the intricacies of plot assessment.

The framework established by \textsf{autovi} and \textsf{autovi.web}
opens up exciting possibilities for further research and development.
Future work could explore the extension of these automated assessment
techniques to other types of diagnostic plots and statistical models,
potentially revolutionising how we approach statistical inference using
visual displays more broadly.

\section{Resources and supplementary
material}\label{resources-and-supplementary-material}

The current version of \textsf{autovi} can be installed from CRAN, and
source code for both packages are available at
\href{https://github.com/TengMCing/autovi}{\texttt{github.com/TengMCing/autovi}}
and
\href{https://github.com/TengMCing/autovi_web}{\texttt{github.com/TengMCing/autovi\_web}}
respectively. The web interface is available from
\href{https://autoviweb.netlify.app/}{\texttt{autoviweb.netlify.app}}.

This paper is reproducibly written using \textsf{Quarto}
\citep{Allaire_Quarto_2024} powered by \textsf{Pandoc}
\citep{MacFarlane_Pandoc} and \textsf{pdfTeX}. The full source code to
reproduce this paper is available at
\href{https://github.com/TengMCing/autovi_paper}{\texttt{github.com/TengMCing/autovi\_paper}}.

These \textsf{R} packages were used for the work: \textsf{tidyverse}
\citep{tidyverse}, \textsf{lmtest} \citep{lmtest}, \textsf{kableExtra}
\citep{kableextra}, \textsf{patchwork} \citep{patchwork},
\textsf{rcartocolor} \citep{rcartocolor}, \textsf{glue} \citep{glue},
\textsf{here} \citep{here}, \textsf{magick} \citep{magick},
\textsf{yardstick} \citep{yardstick} and \textsf{reticulate}
\citep{reticulate}.

  \bibliography{bibliography.bib}

\end{document}